\documentclass{paper}
\usepackage{spconf,amsmath,graphicx}
\usepackage{indentfirst}
\usepackage[pagebackref=true,breaklinks=true,letterpaper=true,colorlinks,bookmarks=false]{hyperref}
\usepackage{mathtools}

\usepackage{amssymb}
\usepackage{subcaption}
\usepackage{multirow}
\usepackage{siunitx}

\usepackage{color, colortbl}
\definecolor{LightCyan}{rgb}{0.88,1,1}
\definecolor{LightRed}{rgb}{1,0.88,0.95}
\usepackage[ruled]{algorithm2e}
\SetAlCapSkip{1em}
\usepackage{comment}
\usepackage{amsmath}
\usepackage{booktabs}

\DeclareMathOperator*{\argmax}{\arg\!\max}
\SetKw{Continue}{continue}
\title{SEMI-SUPERVISED 3D OBJECT DETECTION VIA ADAPTIVE PSEUDO-LABELING}
\name{Hongyi Xu$^{1 \star}$ \thanks{$^{\star}$E-mail: xuhongyi@sjtu.edu.cn}, Fengqi Liu$^{1}$, Qianyu Zhou$^{1}$, Jinkun Hao$^{3}$, Zhijie Cao$^{1}$, Zhengyang Feng$^{1}$, Lizhuang Ma$^{1,2 }$}
  
 \address{$^{1}$Department of Computer Science and Engineering, Shanghai Jiao Tong University, Shanghai, China\\
 $^{2}$School of Computer Science and Technology, East China Normal University, China\\
$^{3}$Department of Automation, East China University of Science and Technology, Shanghai, China
 }

\begin{document}
%\ninept
%
\maketitle
\begin{abstract}
3D object detection is an important task in computer vision. Most existing methods require a large number of high-quality 3D annotations, which are expensive to collect. Especially for outdoor scenes, the problem becomes more severe due to the sparseness of the point cloud and the complexity of urban scenes. Semi-supervised learning is a promising technique to mitigate the data annotation issue. Inspired by this, we propose a novel semi-supervised framework based on pseudo-labeling for outdoor 3D object detection tasks. We design the Adaptive Class Confidence Selection module (ACCS) to generate high-quality pseudo-labels. Besides, we propose Holistic Point Cloud Augmentation (HPCA) for unlabeled data to improve robustness. Experiments on the KITTI benchmark demonstrate the effectiveness of our method. 
% Code and supplementary material are available at \url{https://github.com/tayson0825/SS3DOD}.
% $our method$.

\end{abstract}
\begin{keywords}
3D object detection, point clouds, semi-supervised learning
\end{keywords}
\begin{figure*}[t]
\centering
\setlength{\abovecaptionskip}{2mm}
\includegraphics[scale=0.22]{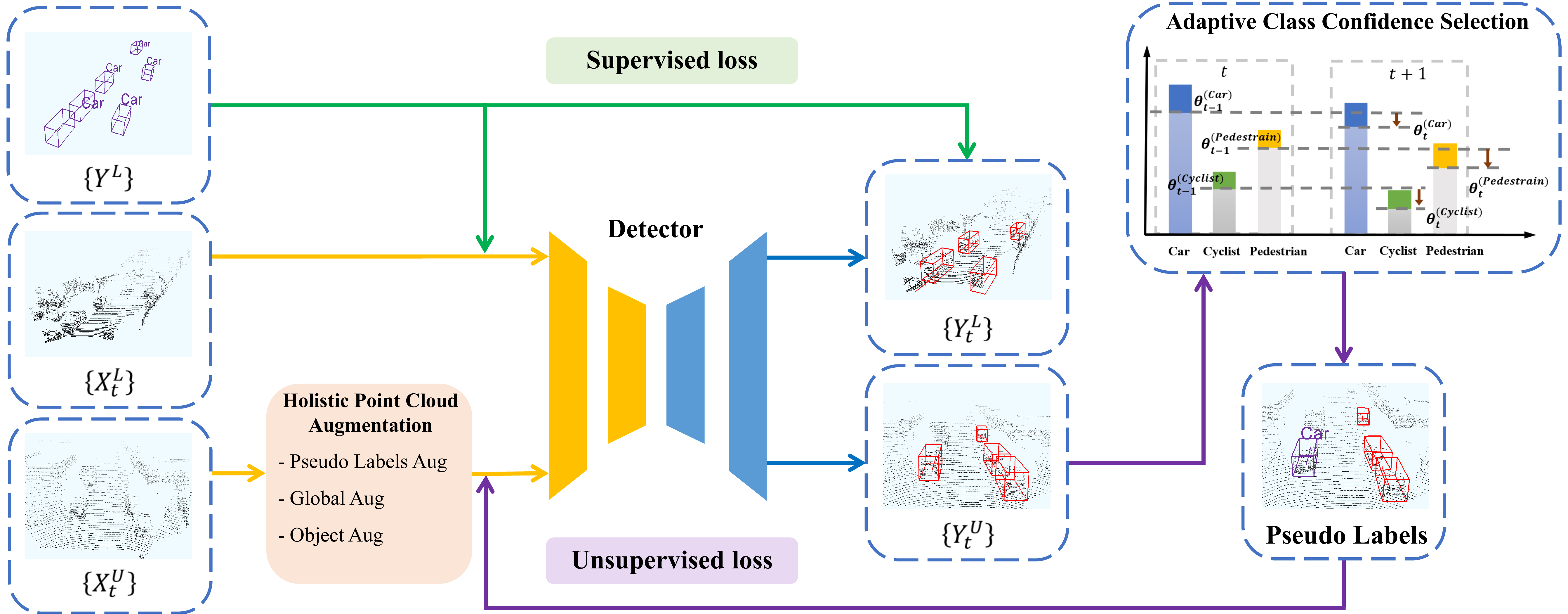}
   \caption{Overview of the proposed semi-supervised 3D object detection framework. We apply strong data augmentation for unlabeled point clouds. Then we put the combination of labeled and unlabeled data into the detector. The supervision for unlabeled data is pseudo labels generated by the ACCS module during the inference iteration.}

\label{fig:Fig1}
\end{figure*}

\vspace{-0.3cm}
\section{Introduction}
\label{sec:intro}
3D object detection~\cite{ding2019votenet,ku2018joint,graham20183d} is an important task in computer vision and has various applications such as autonomous driving~\cite{chen2017multi} and robotics~\cite{fan2020computer,vincent2020dynamic}. The goal of this task is to estimate the categories and corresponding 3D bounding boxes of all targets in the scene. In recent years, great success has been made with the help of deep networks and large-scale datasets
% ~\cite{ding2019votenet,qi2020imvotenet,}. 
~\cite{qi2020imvotenet,yang2018pixor,yan2018second,zhou2018voxelnet}. All of these methods adopt a fully-supervised scheme and require a large number of annotated data, which makes this important task costly. In outdoor scenes, the cost of the annotation is much higher than that of indoor scenes due to the sparseness of the point cloud and the complexity of urban scenes. Therefore, a more effective solution is required to improve the network performance for the outdoor scene without a large quantity of annotated data.

To alleviate the problem of lacking annotated data, semi-supervised learning (SSL)~\cite{rasmus2015semi,samuli2017temporal,miyato2018virtual,feng2020dmt,feng2020semi} and unsupervised domain adaptation (UDA)~\cite{zhou2020uncertainty, zhou2021sad, guo2021label, zhou2021context} brings a feasible solution. SSL allows us to exploit more information from unlabeled data. We can achieve decent performance with only a small amount of labeled data. To the best of our knowledge,~\cite{zhao2020sess} is the only deep learning method for 3D object detection under semi-supervised conditions. Promising results have been achieved on indoor scenes~\cite{sunrgb-d,scannet}. However, the consistency strategy they adopted may face challenges in those more sparse outdoor point cloud scenes. 

Motivated by these facts, in this work, we propose a novel semi-supervised 3D object detection method based on pseudo-labeling for outdoor  scenes.Considering that the direct use of pseudo-labeling~\cite{iscen2019label} will produce inevitable noise that may accumulate errors and affect the performance, we adopt Adaptive Class Confidence Selection scheme (ACCS) for pseudo-label filtering. Specifically, we present ACCS by sorting predicted confidence and gradually increasing the ratio of pseudo labels. With the adaptive selection, we can greatly alleviate the over-confident errors of pseudo-labeling.
Besides, we design Holistic Point Cloud Augmentation (HPCA) to unlabeled data to improve robustness by applies a combination of strong data augmentations, which includes pseudo label augmentation, global augmentation and object augmentation. These augmentations on unlabeled point clouds can further improve the robustness of the neural networks.

Our main contributions can be summarized as follows:
% \vspace{-0.25cm}
\begin{itemize}\setlength{\itemsep}{0pt}
\item  We propose a novel semi-supervised 3D object detection framework via adaptive pseudo-labeling for outdoor 3D scenes. The effectiveness is validated on KITTI benchmark.
\item We present Adaptive Class Confidence Selection scheme (ACCS) to adaptively implement pseudo-label filtering on each class.
\item We design Holistic Point Cloud Augmentation scheme (HPCA) to regularize unlabeled data and the detector in each iteration of pseudo-labeling.
\end{itemize}

\vspace{-0.5cm}
\section{PROPOSED METHOD}
\label{sec:pagestyle}
\subsection{Preliminaries}
Our method directly uses the original point cloud as input. We divide our detection goal into classify objects in the 3D scene and locate their 3D bounding boxes. Besides, the overall structure is illustrated in Fig.~\ref{fig:Fig1}. For data samples, we have $N$ training samples, including  
$N_{l}$ labeled point clouds 

$\mathcal{P}^{U}=\left\{{X}^{U}\right\}$, ${X}^{U}=({x}_{1}^{U},...,{x}_{i}^{U})$. Each object in $\mathbf{x}_{i}^{L} \in \mathbb{R}^{n \times 3}$ is labeled by a semantic class $k$ ($K$ predefined classes in total)  represents the point cloud of the 3D scene, including $n$ coordinate points; $\mathbf{y}_{i}^{L}$ represents the ground truth annotation corresponding to $\mathbf{x}_{i}^{L}$. $t \in\{1, \cdots, T\}$ is the number of iterations. We define our 3D detection network as $f_{\theta}^{t} $, where $\theta$ are the model parameters. 3D bounding boxes are parameterized by $(x_{c}, y_{c}, z_{c}, l, w, h, \delta  )$ in LiDAR coordinate system, where $\left( x_{c}, y_{c}, z_{c} \right)$ is the center position, $\left( l, w, h \right)$ is the  size, and $\delta $ is the object orientation along the vertical  $z$ axis.
% zhijie: 行内公式应该包含在$$中 
\subsection{Adaptive Class Confidence Selection}
The fundamental assumption of pseudo-labeling is that the network can understand more data structure if given high confidence prediction. To this end, we can take those unlabeled data which has high confidence prediction as labeled data in the next iteration. However,if we choose the same confidence filtering threshold for each class in each iteration, it is inevitable to result in an inter-class imbalance where result in the model bias towards easy classes since they have higher prediction confidence than hard classes.

Therefore, we design the adaptive class confidence selection module (ACCS), which guarantees that different classes should have different thresholds. Specifically, ACCS module selects the confidence of the samples with a pre-set top ratio $\mathbf{c}$ in $Y_t^U$ predicted by model $f_{\theta}^{t} $ as the threshold $\lambda_{t}^{k}$ for pseudo-labeling. The initial ratio $\mathbf{c}$ for classes is set the same. Classes can adaptively adjust the confidence threshold based on ratio $\mathbf{c}$ according to different class difficulty (easy or hard). In addition, as the performance of the model increases with the iterations of pseudo-labeling, we increase the ratio $c$ in each round gradually to ensure our model learns more data structure information. Meanwhile, the confidence threshold for each class can also dynamically change, allowing more diverse samples to be selected. This can be formulated as follows:
\begin{equation}
    \mathbf{c}_{t+1} = \mathbf{c}_{t} + \Delta\mathbf{c},
\end{equation}
where $\mathbf{c}_{t}$ and $\mathbf{c}_{t+1}$ represent the ratio in $t$  and $t+1$ iteration respectively. $\Delta\mathbf{c}$ represents the increasement of pseudo-labeling ratio between consecutive iteration.

With ACCS, our adaptive pseudo-labeling strategy can be more effective. Algorithm~\ref{alg:al1} shows the process. We first use the labeled data $\mathcal{P}^{L}$ to train the initial detection model $f_{\theta}^{0}$. In the next $T$ iterations, we use Holistic Point Cloud Augmentation (HPCA, Section \ref{ssec:data aug}) to enhance all unlabeled data $\mathcal{P}^{U}$. Then the current model $f_{\theta}^{t} $ produces the prediction results  $Y_{t}^{U}$ of $\mathcal{P}^{U}$. After that, we use the proposed ACCS to select the confidence of the sample with top ratio $c$ in $Y_{t}^{U}$ as the threshold of pseudo label. The  Percentile $(X, c)$ returns the confidence value of the $c$-th percentile ratio. Besides, samples in $X_t^{U}$ will be added to the training sample set $X_{t}$ if their confidence is larger than the threshold $\lambda_{t}^{k}$. We use new sample set $(X_{t},Y_{t})$ to retrain our detection network, and start a new iteration until $T$ round. Finally, we predict the label of all the data.

Let $k \leftarrow \argmax_{k}\{\frac{p(k|{\mathbf{x}_{t}^{U}};f_{\theta}^{t})}{\lambda_t^{k}}\}$ be the predicted class, denote pseudo label selection result in round $t$ by binary weight ${m}_{t}^{U}$ ($1$ corresponds to selected 3D bounding box, $0$ ignored):
\begin{equation}\label{eq1}
{m}_{t}^{U}=\left\{
\begin{aligned}
1, &~\text{if}~p(k|{\mathbf{x}_{t}^{U}};f_{\theta}^{t})>{\lambda_t^{k}}\\
0, &~\mathrm{otherwise}
\end{aligned}
\right.
\end{equation}
 
Our supervised 3D bounding box regression loss follows PointRCNN~\cite{shi2019PointRCNN}, and can be expressed as:
\begin{small}
\begin{equation}\label{eq2}
\mathcal{L}_{s} = \frac{1}{N_{\mathrm{pos}}} \sum_{p \in \mathrm{pos}}\left(\mathcal{L}_{\mathrm{bin}}(\mathbf{x}_{p}^{L}, \mathbf{y}_{p}^{L}) + \mathcal{L}_{\mathrm{res}}(\mathbf{x}_{p}^{L}, \mathbf{y}_{p}^{L})\right),
\end{equation}
\end{small}
where $\mathcal{L}_{\mathrm{bin}}$ and $\mathcal{L}_{\mathrm{res}}$ are the two losses of the two-stage pipeline. According to Eq.~(\ref{eq1}), the unsupervised loss $\mathcal{L}_{u}$ is:

\begin{footnotesize}
\begin{equation}
\mathcal{L}_{u} = \frac{1}{N_{\mathrm{pos}}} \sum_{p \in \mathrm{pos}}{m}^{U}\left(\mathcal{L}_{\mathrm{bin}}\big(\mathcal{A}(\mathbf{x}_{p}^{U}, \hat{\mathbf{y}}_{p}^{U})\big) + \mathcal{L}_{\mathrm{res}}\big(\mathcal{A}(\mathbf{x}_{p}^{U}, \hat{\mathbf{y}}_{p}^{U})\big)\right),
\end{equation}
\end{footnotesize}
where $m^{U}$ is the similar binary weight of the corresponding 3D point cloud box of $\mathbf{x}_{p}^{U}$. $\mathcal{A}(\cdot)$ is the HPCA scheme (Section \ref{ssec:data aug}).

Therefore, our overall loss is $\mathcal{L} = \mathcal{L}_{s} + \mathcal{L}_{u}$.

\begin{algorithm}[htb]
\footnotesize
\caption{Adaptive Pseudo-Labeling Strategy}\label{alg:al1}
\SetNoFillComment
\DontPrintSemicolon
\SetInd{0.1em}{0.7em}
\SetAlgoLined
\textbf{Input}: {$\mathcal{P}^{L}$}: labeled point cloud samples;{$\mathcal{P}^{U}$}: unlabeled point cloud samples

\textbf{Require}: {$f_{\theta}^{t}$}: detection model in round t; {$K$}: number of class; {$\lambda_{k}$}: Threshold for each class; {$t \in\{1, \cdots, T\}$}: number of iterations; $\mathbf{c}$: initial confidence ratio in ACCS; $\Delta\mathbf{c}$: the growth of $\mathbf{c}$ per round in pseudo-labeling

\textbf{Output}: final precdited label ${Y}^{all}$ for all  data\\

\textbf{Procedure}\\

 $X_{t} := \mathcal{P}^{L}$\\

{$f_{\theta}^{t}$} $\leftarrow$ train using $(X_{t},Y^{L})$ only\\

\For{t = 1 {\rm to} T}{ 
 $ X_t^{U} := \mathcal{P}^{U}$\\
%$ X_t^{U} $\leftarrow$ Subset(\mathcal{P}^{U})$\\
 $Y_{t}^{U}$  $\leftarrow$  {$f_{\theta}^{t}(X_t^{U})$} \\
\textbf{ ACCS:} ${\lambda_{k}}=\textit{Percentile}(sort(Y_{t}^{U},order=descending),ratio=\mathbf{c}) $, $k\in K$\\
$\mathbf{c} := \mathbf{c}+\Delta\mathbf{c}$

\For{$ x_t^{U} \in X_t^{U}, y_t^{U} \in Y_t^{U}$}{
  \If {$y_{t}^{U}$ $>$ ${\lambda_{k}}$} {
    $X_{t} := X_{t} \cup x_t^{U}$
    $Y_{t} := Y_{t} \cup y_t^{U}$
    }
}

{$f_{\theta}^{t}$} $\leftarrow$ train with$(X_{t},Y_{t})$ using \textbf{HPCA} for unlabeled data\\
$t := t + 1$\\
}
${Y}^{all}$ $\leftarrow$ Use {$f_{\theta}^{T}$}  to infer all data\\
return ${Y}^{all}$\\
\textbf{end}\\
\end{algorithm}
\vspace{-0.5cm}

\begin{table*}[]
\small
%\captionsetup{font=small}
\renewcommand{\arraystretch}{0.95}
\centering
\begin{tabular}{c c ccc ccc ccc}
\toprule
\multirow{2}{*}{Ratios} & \multirow{2}{*}{Model} & \multicolumn{3}{c}{Car (IoU=0.7)}                         & \multicolumn{3}{c}{Pedestrain (IoU=0.5)}                  & \multicolumn{3}{c}{Cyclist (IoU=0.5)}                     \\ \cmidrule(lr){3-5} \cmidrule(lr){6-8}  \cmidrule(lr){9-11}
                            &                        & Easy           & Moderate         & Hard           & Easy           & Moderate         & Hard           & Easy           & Moderate         & Hard           \\ \midrule
\multirow{3}{*}{10\%}    & PointRCNN           & 52.20           & 40.29          & 38.44          & 34.30           & 30.22          & 26.38          & 46.79          & 30.83          & 28.55          \\
                            & \textbf{Ours}          & \textbf{70.32} & \textbf{57.29} & \textbf{52.13} & \textbf{49.59} & \textbf{43.25} & \textbf{38.69} & \textbf{76.68} & \textbf{53.36} & \textbf{51.26} \\
                            & Improv.(\%)          & 34.73$\uparrow$          & 42.20$\uparrow$          & 35.61$\uparrow$          & 44.59$\uparrow$          & 43.09$\uparrow$          & 46.66$\uparrow$          & 63.89$\uparrow$          & 73.10$\uparrow$          & 79.58$\uparrow$          \\ \midrule
\multirow{3}{*}{20\%}    & PointRCNN           & 57.79          & 47.54          & 42.33          & 42.70           & 37.24          & 30.59          & 47.94          & 33.50           & 30.96          \\
                            & \textbf{Ours}          & \textbf{72.19} & \textbf{61.36} & \textbf{54.29} & \textbf{49.76} & \textbf{44.61} & \textbf{38.95} & \textbf{77.26} & \textbf{53.84} & \textbf{51.92} \\
                            & Improv.(\%)           & 24.92$\uparrow$   & 29.07$\uparrow$ & 28.25$\uparrow$ & 16.54$\uparrow$ & 19.79$\uparrow$ & 27.34$\uparrow$ & 61.17$\uparrow$ & 60.71$\uparrow$          & 67.69$\uparrow$          \\ \midrule
\multirow{3}{*}{30\%}    & PointRCNN           & 71.81          & 58.37          & 50.66          & 48.30           & 43.27          & 39.47          & 54.7           & 37.77          & 35.1           \\
                            & \textbf{Ours}          & \textbf{75.42} & \textbf{63.26} & \textbf{58.07} & \textbf{50.93} & \textbf{44.64} & \textbf{40.99} & \textbf{77.6}  & \textbf{55.61} & \textbf{53.08} \\
                            & Improv.(\%)           & 5.04$\uparrow$           & 8.37$\uparrow$           & 14.62$\uparrow$          & 5.44$\uparrow$         & 3.16$\uparrow$           & 3.86$\uparrow$           & 41.87$\uparrow$          & 47.24$\uparrow$          & 51.24$\uparrow$          \\ \midrule
\multirow{3}{*}{50\%}    & PointRCNN           & 77.54          & 62.72          & 54.65          & 50.29          & 43.57          & 39.94          & 55.06          & 38.58          & 36.97          \\
                            & \textbf{Ours}          & \textbf{77.94} & \textbf{64.19} & \textbf{58.72} & \textbf{50.94} & \textbf{45.42} & \textbf{41.27} & \textbf{78.43} & \textbf{55.62} & \textbf{54.81} \\
                            & Improv.(\%)          & 0.52$\uparrow$           & 2.34$\uparrow$           & 7.45$\uparrow$           & 1.30$\uparrow$           & 4.23$\uparrow$           & 3.32$\uparrow$           & 42.44$\uparrow$          & 44.15$\uparrow$          & 48.26$\uparrow$          \\ \bottomrule

\end{tabular}
  \vspace{-2mm}
  \caption{Comparsion with PointRCNN~\cite{shi2019PointRCNN} on KITTI~\cite{geiger2012we} val split with different ratios of labeled data. The evaluation metric is Average Precision (AP) with IoU threshold 0.7 for car and 0.5 for pedestrian/cyclist as ~\cite{shi2019PointRCNN}. Results are the mean of 3 random runs.}
  \vspace{-2mm}
\label{table:111}
\end{table*}
\subsection{Holistic Point Cloud Augmentation (HPCA)}\label{subsec:hpca}
\label{ssec:data aug}

As mentioned in ~\cite{zhao2020sess,sohn2020simple,zhou2020uncertainty}, applying a strong data augmentation is very important for semi-supervised approaches. The data augmentation scheme of supervised and semi-supervised strategies for 2D classification tasks has been extensively studied in~\cite{berthelot2019remixmatch,cubuk2019autoaugment,cubuk2019randaugment}. However, little effort has been made on 3D outdoor object detection. Consequently, we propose Holistic Point Cloud Augmentation (HPCA) which applies strong data augmentations on unlabeled point cloud data. Through adding perturbation and noise to the unlabeled point clouds, we can improve the generalization of the model, and prevent learning from unintended properties, such as the order and absolute location of points.

\noindent\textbf{Pseudo Labels Augmentation.} The major obstacle in semi-supervised 3D object detection is the existence of few ground truths in each point cloud scene, which slows convergence and hinders the performance of the detector. To solve this problem and simulate various and complex environments, inspired by~\cite{yan2018second}, we introduce a strong data augmentation which mixes pseudo detection labels between different point cloud scenes. Firstly, we generate a ground-truth database comprising high confidence pseudo labels in the last pseudo-labeling iteration and corresponding points contained in the 3D bounding boxes. Then, we randomly sample several pseudo labels and points inside with non-overlapping bounding boxes from the database before each training step, which can greatly increase the number of ground truths in each point cloud.

\noindent\textbf{Global Augmentation.} After pseudo labels augmentation, we use stochastic transformation operations sequentially. These operations are globally applied to the entire input point cloud. We utilize random scaling, random rotation along the vertical $z$ axis, and random horizontal flip along $x$ axis in the LiDAR coordinate system. The augmentation operations sequence can be formulated as $\mathcal{A}_{global} = \left\{ s, R, f  \right\}$. Specifically, $s$ is the scale factor drawn from a uniform distribution $\left[ a,b\right]$; $R$ represents the rotation matrix with the rotation angle $\varphi$ sampled from a uniform distribution $\left[ -\phi,+\phi \right]$; $f$ is the indicator $\eta$ uniformly drawn from $[0,1]$, which indicates whether to flip.

\noindent\textbf{Object Augmentation.} Object data augmentation applies random noise separately to each pseudo-labeled 3D bounding box and points within the box. Similar to the global object augmentation, the object augmentation sequence can be formulated as $\mathcal{A}_{object} = \left\{ G, T  \right\}$. Inspired by VoxelNet~\cite{zhou2018voxelnet}, we use random rotation by matrix $G$ with a rotation angle $\Delta \theta \sim \mathbf{U}(-\psi,+\psi)$ and random linear translation by matrix $T = (\Delta x,\Delta y,\Delta z)$, where $(\Delta x,\Delta y,\Delta z)$ is sampled from a standard Gaussian distribution.

% \vspace{-0.45cm}
\section{EXPERIMENTAL RESULTS}
\label{sec:typestyle}

\subsection{Experiment Setup}
\noindent\textbf{Datasets.}
We perform extensive experiments on the popular 3D object detection benchmark KITTI~\cite{geiger2012we}, which is comprised of 7,481 training images and 7,518 test images as well as the corresponding point clouds. We further divide the training samples into training and validation split following~\cite{chen2017multi}.

\begin{figure}[!h]
\centering
\setlength{\abovecaptionskip}{0.cm}
\includegraphics[scale=0.155]{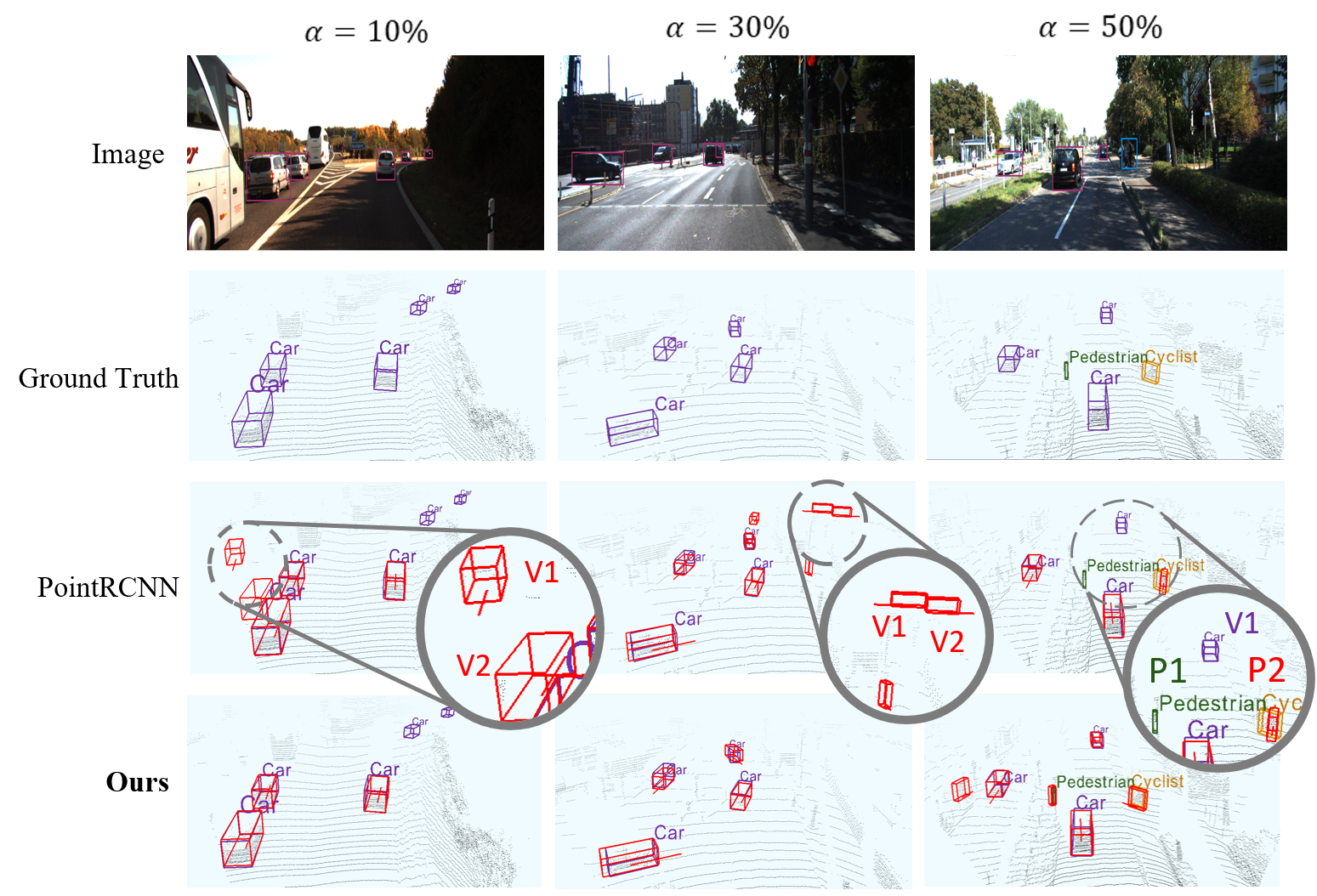}
\caption{Qualitative results and baseline on KITTI~\cite{geiger2012we} val split(The number of class is 3). Camera image and point cloud scene with ground truth (row 1, 2), 3D object detection result (row 3, 4). $\alpha$ is the ratio of labeled data. Each column shows the detection results of 3D objects at different ratio $\alpha$. The ground-truth boxes for cars, pedestrians, and cyclists are purple, green, and yellow respectively. The prediction boxes are red. Note that ``V'' and ``P'' highlighted in the orange circles indicate cars and people (including pedestrians and cyclists) which are falsely detected by other methods.}
\label{fig:Fig_2}
\end{figure}

\noindent\textbf{Settings.}
Firstly, we divide the training set of KITTI into several sequences where point clouds come from the same video. Then, we randomly sample the training set with 10\%, 20\%, 30\%, and 50\% labeled data according to the sequence, which can ensure that labeled data $\mathcal{P}^{L}$ and unlabeled data $\mathcal{P}^{U}$ are derived from different videos to the greatest extent.

\noindent\textbf{Implementation Details.}
We adopt PointRCNN~\cite{shi2019PointRCNN} as the 3D object detection network. The input of our network is divided into labeled and unlabeled data. For each point cloud scene in the training set, we subsample 16,384 points. In a batch, we randomly sample $B_{l}$ samples with the label from $\mathcal{P}^{L}$ and $B_{u}$ samples without label from $\mathcal{P}^{U}$. In our experiment, the ratio of $B_{l}$ to $B_{u}$ is 1:1. Here we set the number of iterations $T=5$, the initial confidence ratio $\mathbf{c}=20\%$ and increment ratio $\Delta\mathbf{c}=5\%$.

\subsection{Evaluation and Discussion}

In this section, we compare our semi-supervised method with the baseline method PointRCNN. Note that the baseline method is trained on $\mathcal{P}^{L}$ with different ratios in a fully supervised manner.

\noindent\textbf{Quantitative Analysis.}
Table~\ref{table:111} shows that the method based on pseudo-labeling and data augmentation we adopted can significantly outperform PointRCNN under all label ratio settings. As the proportion of labeled samples decreases, the performance gap between our method and fully supervised PointRCNN becomes wider. Especially when 10\% of label data is given, our method is improved by about 24.30\% relative to PointRCNN. This proves our method can learn more useful information from unlabeled data to help improve detection accuracy. Our framework can also be integrated with any other supervised 3D object detector.\\
\noindent\textbf{Qualitative Analysis.}
We present the 3D object detection results at different ratios in Fig.~\ref{fig:Fig_2}. Results show that our method is better than PointRCNN in every ratio setting. The first column indicates that we have less false detection in the case of occlusion. By contrast, PointRCNN wrongly detects cars ``V1'' and ``V2''. The second column shows that our framework has fewer errors in the detection of distant sparse points. The third row demonstrates that our framework is better at detecting multiple categories of 3D objects and can estimate the orientation of the objects more accurately. By contrast, PointRCNN fails to detect pedestrian ``P1'', car ``V1'' and misjudges the orientation of cyclist ``P2''.

% \vspace{-1.2cm}
\section{Conclusion}
\label{sec:majhead}
% \vspace{-0.53cm}
We propose a novel semi-supervised 3D object detection framework based on pseudo-labeling for outdoor scenes that only utilize few labeled data to achieve promising performance. Our Adaptive  Class  Confidence  Selection module (ACCS) adaptively implements pseudo-label filtering for each class and greatly alleviates the problem of class imbalance. Besides, Holistic Point Cloud Augmentation (HPCA) regularizes unlabeled data in each iteration of pseudo-labeling. Experimental results show that our proposed method significantly outperforms the existing baseline on KITTI benchmark, which demonstrates the effectiveness of our work.

\noindent
\textbf{Acknowledgement.}
% \hspace{0.3cm}
This work is sponsored by National Natural Science Foundation of China (61972157), National Key Research and Development Program of China (2019YFC1 521104), Shanghai Municipal Science and Technology Major Project (2021SHZDZX0102).

% -------------------------------------------------------------------------
\bibliographystyle{ieee}
\bibliography{ref}

\end{document}